\documentclass[sigconf,nonacm]{acmart}

\AtBeginDocument{%
  \providecommand\BibTeX{{%
    \normalfont B\kern-0.5em{\scshape i\kern-0.25em b}\kern-0.8em\TeX}}}

\renewcommand\footnotetextcopyrightpermission[1]{}

\usepackage{array,booktabs}
\usepackage{bigints}
\usepackage{algorithm,algorithmic,xcolor}

\graphicspath{{images/}}

\definecolor{plum}{rgb}{0.45,0,.66}
\definecolor{mythistle}{rgb}{.99,.195,.133}
\definecolor{myred}{cmyk}{0.000000,1.000000,1.000000,0.1}
\definecolor{myblue}{cmyk}{1.000000,0.750000,0.000000,0.1}
\definecolor{mybgn}{cmyk}{0.850000,0.350000,0.000000,0.1}
\definecolor{mygrn}{cmyk}{0.750000,0.000000,1.000000,0.2}

\newcommand{\bl}{\begin{itemize}}
\newcommand{\el}{\end{itemize}}
\newcommand{\be}{\begin{enumerate}}
\newcommand{\ee}{\end{enumerate}}

\newcommand{\bea}{\begin{eqnarray*}}
\newcommand{\eea}{\end{eqnarray*}}

\newcommand{\beq}{\begin{equation}}
\newcommand{\eeq}{\end{equation}}

\newcommand{\bmx}{\left[ \begin{array}}
\newcommand{\emx}{\end{array} \right]}

\def\half{\frac{1}{2}}
\def\flip{\hat{\bf{x}}^c}

\newcommand{\mybf}[1]{\textbf{\em #1} }

\def\bfx{{\mybf{x}}}

\def\bfz{{\mybf{z}}}

\def\bfB{{\mybf{B}}}

\def\bfD{{\mybf{D}}}

\def\bfF{{\mybf{F}}}

\def\bfP{{\mybf{P}}}
\def\bfQ{{\mybf{Q}}}
\def\bfR{{\mybf{R}}}

\DeclareMathOperator{\softmax}{softmax}

\def\bfD{\boldsymbol{D}}

\newcommand{\dy}[1]{#1}
\newcommand{\dd}[1]{#1}
\newcommand{\da}[1]{#1}
\newcommand{\dc}[1]{#1}

\newcommand{\ra}[1]{#1}
\newcommand{\rb}[1]{#1}

\begin{document}

\title{Auditing and Debugging Deep Learning Models via Decision Boundaries\ra{: Individual-level and Group-level Analysis}}

\author{Roozbeh Yousefzadeh}
\affiliation{%
  \institution{Departments of Genetics and Computer Science}
  \institution{Yale University}
}
\email{roozbeh.yousefzadeh@yale.edu}

\author{Dianne P. O'Leary}
\affiliation{%
  \institution{Department of Computer Science}
  \institution{and Institute for Advanced Computer Studies}
  \institution{University of Maryland, College Park}
}
\email{oleary@cs.umd.edu}

%
%
%
%
%
%

\renewcommand{\shortauthors}{Yousefzadeh and O'Leary}

\begin{abstract}
Deep learning models have been criticized for their lack of easy interpretation, which undermines confidence in their use for important applications. Nevertheless, they are consistently utilized in many applications, consequential to humans' lives, mostly because of their better performance. Therefore, there is a great need for computational methods that can explain, audit, and debug such models. Here, we use {\em flip points} to accomplish these goals for deep learning models \da{with continuous output scores (e.g., computed by softmax)}\ra{, used in social applications}.
A flip point is any point that lies on the boundary between two output classes: e.g. for a model with a binary yes/no output, a flip point is any input that generates equal scores for ``yes" and ``no". The flip point closest to a given input is of particular importance because it reveals the least changes in the input that would change a model's classification, and we show that it is the solution to a well-posed optimization problem. Flip points also enable us to systematically study the decision boundaries of a deep learning classifier. The resulting insight into the decision boundaries of a deep model can clearly explain the \ra{model's output on the individual-level,} via an explanation report that is understandable by non-experts. \ra{We also develop a procedure to understand and audit model behavior towards groups of people. Flip points can also be used to alter the decision boundaries in order to improve undesirable behaviors. We demonstrate our methods by investigating several models trained on standard datasets used in social applications of machine learning. }We also identify the features that are most responsible for particular classifications and misclassifications.
\end{abstract}




\maketitle


\section{Introduction}

\ra{Our focus in this paper is auditing and debugging deep learning models in social applications of machine learning.}
In these applications, deep learning models are usually trained for a specific task and then used, for example to make decisions or to make predictions. Despite their unprecedented success in performing machine learning tasks accurately and fast, these trained models are often described as black-boxes because they are so complex that their output is not easily explainable in terms of their inputs. \da{As a result, in many cases, no explanation of decisions based on these models can be provided to those affected by them \citep{zeng2017interpretable}.}

\begin{figure}[t]
\centering
\includegraphics[width=1\columnwidth]{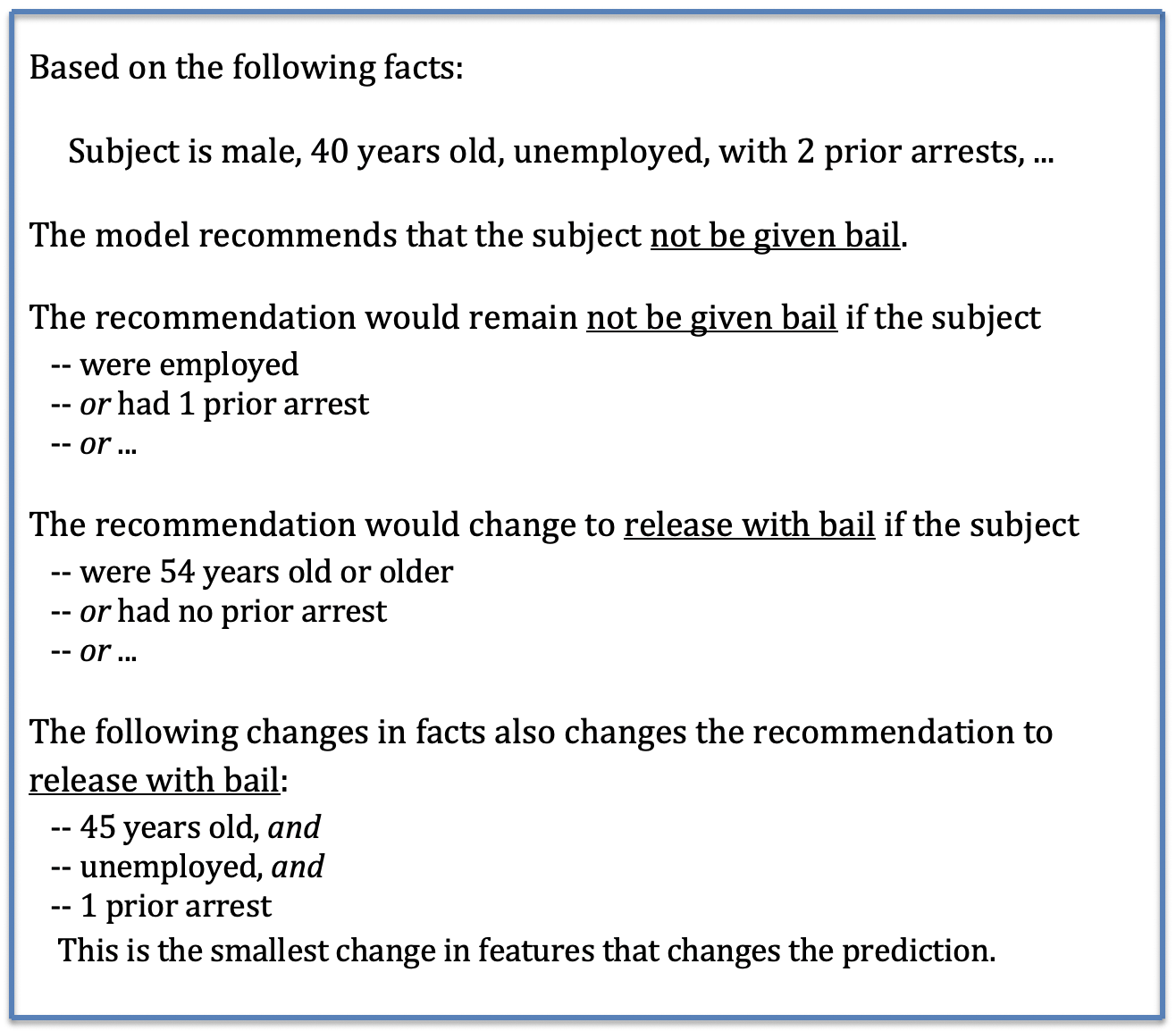}
\caption{Example of the kind of information that can be obtained by calculating flip points. We answer questions such as, ``For a particular input to a deep learning model, what is the smallest change in a single continuous feature that changes the output of the model? What is the smallest change in a particular set of features that changes the output?"} 
\label{fig_explain_report}
\end{figure}

This inexplainability becomes problematic when deep learning models are utilized in tasks consequential to human lives, such as in criminal justice, medicine, and business. Independent studies have revealed that many of these black-box models have unacceptable behavior, for example towards features such as race, age, etc. of individuals \citep{rudin2018age}. Because of this, there have been calls for avoiding deep learning models in high-stakes decision making \cite{rudin2019stop}. Additionally, laws and regulations have been proposed to require decisions made by the machine learning models to be accompanied with clear explanations for the individuals affected by the decisions \cite{wachter2018reasonable}. Several methods have been \ra{developed} to explain the outputs of \da{models simpler than deep learning models to non-expert users such as administrators or clinicians \citep{zeng2017interpretable,rudin2018learning,narayanan2018humans,lage2019evaluation}.
In contrast, existing} interpretation methods for deep learning models either lack the ability to directly communicate with non-expert users or have limitations in their scope, computational ability, or accuracy, as we will explain in the next section.

\ra{In the meantime, d}eep learning is ever more widely used on important applications in order to achieve high accuracy, scalability, etc. \ra{Sometimes, deep learning models are utilized even when they do not have a clear advantage over simple models, perhaps to avoid transparency or to preserve the models as proprietary \citep{rudin2019why}. While it is not easy to draw the line as to where their use is advantageous, it is important to have the computational tools to thoroughly audit the models, provide the required explanations for their outputs, and/or to expose their flaws and biases. It would also be useful to have the tools to change their undesirable behavior.}


\ra{We provide tools for two levels of auditing: {\em individual-level} and {\em group-level}.} The type of feedback that our methods provide \ra{on the individual-level} is illustrated in Figure \ref{fig_explain_report} and discussed in Section \ref{sec_feedback}; \dy{in particular, we identify sets of features that have no effect on the model's decision and sets that change the decision, and we find the closest input with a different decision.} 
\ra{For group-level analysis, we develop methods to audit the behavior of models towards groups of individuals, for example, people with certain race or certain education.}



In Section \ref{sec_literature}, we review the literature and explain the advantages of our method \dc{compared} to other popular methods such as LIME \citep{ribeiro2016should}.
In Section~\ref{sec_using}, we present our computational approach to perform the above tasks, based on investigating and altering the decision boundaries of deep learning models by computing {\em flip points}, certain interesting points on those boundaries, defined in Section~\ref{sec_define_flip}, \da{where we also introduce the concept of constrained flip points}. In Section~\ref{sect_results}, we present our numerical results on three different datasets with societal context. Section~\ref{sec_compare} compares our methods with other applicable methods in the literature. Finally, in Section~\ref{sec_conclusion}, we present our conclusions and directions for future work.


\section{Literature review} \label{sec_literature}

There have been several approaches proposed for interpreting deep learning models and other black-box models. \da{Here we  mention} a few papers representative of the field.

\citet{spangher2018actionable} have (independently) defined \textit{a flip set} as the set of changes in the input that can flip the prediction of a classifier. However, their \da{method} is only applicable to linear classifiers such as linear regression models and logistic regression. They use flip sets to explain the least changes in individual inputs but do not go further to interpret the overall behavior of the model or to debug it.

\citet{wachter2018counterfactual} \da{define} \textit{counterfactuals} as the possible changes in the input that can produce a different output label and use them to explain the decision of a model. However,  \da{their closest counterfactual is mathematically ill-defined; for  deep learning models with continuous output, there is no "closest point" with a different output label because there are points arbitrarily close to the decision boundary.}
Moreover, their proposed algorithm uses enumeration, applicable only to a small number of features. \citet{russell2019efficient} later suggested integer programming to solve such optimization problems, but the models used as examples are linear with small dimensionality, and the closest counterfactual in their formulation is ill-defined. 

Some studies have taken a model-agnostic approach to interpreting black-box models. For example, the approach taken by \citet{ribeiro2016should}, known as LIME, randomly perturbs an input until it obtains points on two sides of a decision boundary and then performs linear regression to estimate the location of the boundary in that vicinity. The simplifying assumption to approximate the decision boundary 
\da{with} hyperplanes can be misleading for \ra{deep learning} models, as shown by \cite{fawzi2018empirical,yousefzadeh2019investigating}. Hence, the output of the LIME model and its corresponding explanation may actually contradict the output of the original model, as empirically shown by \cite{white2019measurable}. Another issue in LIME's approach is the reliance on random perturbations of inputs, which has computational limitations. \citet{lakkaraju2019faithful} have also shown via surveys that such explanations may not be effective in communicating with non-expert users. \da{Our} method has an accuracy advantage over LIME, because we find a point exactly on the decision boundary instead of estimating its location via a surrogate linear regression model. Additionally, our explanation report can directly communicate with non-expert users \da{such as credit applicants or clinicians}.

There are approaches that create rule-lists based on the classifications of a deep learning model, and then use the obtained rules to explain the outputs \cite{anchors:aaai18,lakkaraju2017interpretable,lakkaraju2019faithful}. These approaches have serious limitations in terms of scalability and accuracy, mostly because a deep learning model is usually too complex to be emulated via a simple set of if-then rules.
For example, the outputs of the if-then rules obtained by \cite{lakkaraju2019faithful} are different than the outputs of their neural network for more than 10\% of the data points, \da{even though} the feature space has only 7 dimensions. The computation time to obtain the rule-list is also in the order of few hours for the \da{7-feature model}, while we provide the explanation report for an input with 88 features in a few seconds.

\citet{koh2017understanding} and \citet{koh2019accuracy} have used influence functions to reveal the importance of individual training data in forming the trained model, but their method cannot be used to explain outputs of the models or to investigate the decision boundaries.



There are studies in deep learning that consider the decision boundaries from other perspectives. For example, \citet{elsayed2018large} and \citet{marginbased2019} use first-order Taylor series approximation to estimate the distance to decision boundaries for individual inputs, and study the distance in relation to generalization error in deep learning. However, those approximation methods \da{have} been shown to be unreliable for nonlinear models \cite{yousefzadeh2019investigating}. Methods to generate adversarial inputs, for example \citet{fawzi2017robustness,jetley2018friends,moosavi2016deepfool}, apply small perturbations to an input until its classification changes, but those methods do not seek the closest point on the decision boundaries. and therefore cannot find the least changes required to change the model's output. Most recent methods for computing adversarial inputs, such as \citet{ilyas2019adversarial} and \citet{tsipras2018robustness}, also do not \ra{seek points on or near the decision boundaries.}

\section{Defining and computing flip points} \label{sec_define_flip}

\da{
For ease of exposition, in this section we \dc{consider} a model with two continuous outputs. Extensions to models with multi-class outputs or quantified output is straightforward. We first review the work on flip points in \cite{yousefzadeh2019interpreting} and then define constrained flip points.

\subsection{Flip points} \label{sec_define_flipa}

Consider a model $\mathcal{N}$ that has two continuous outputs $z_1$ and $z_2$. For convenience, we assume that they are normalized to sum to 1 (e.g., by softmax) and write $\bfz = \mathcal{N}(\bfx)$.
An output with $z_1(\bfx) > \half$ corresponds to one class, for example, ``cancerous".
Similarly,  $z_1(\bfx) < \half$ might be a prediction of ``noncancerous", and the prediction for $z_1(\bfx) = \half$ is undefined.

We refer to points on the decision boundary $z_1(\bfx) = \half$ as {\em flip points}, and we are particularly interested in the smallest change in a given input $\bfx$ that changes the decision of the model.
We can find this closest flip point $\flip$ by solving an optimization problem 
\begin{eqnarray*} \label{eq_opt}
    \min\limits_{\hat{\bfx} } \| {\hat{\bfx}} - \bfx \| , \\ 
     z_1(\hat{\bfx} )= 1/2.
\end{eqnarray*}
where $\| . \|$ is a norm appropriate to the data.
Specific problems might require additional constraints, for example, upper and lower bounds on image data, or integer constraints on features such as gender.
It is possible that the solution $\flip$ is not unique, but the minimal distance is always unique.

Our optimization problem can be solved by  off-the-shelf or specialized algorithms that determine local minimizers for nonconvex problems. For a neural network, the cost of each iteration in determining a flip point is less than the marginal cost of including one point in one iteration of training the model. For the examples we provide in this paper, computing a closest flip point just takes less than a second on a 2017 Macbook.

Another way of looking at the cost is to observe that the cost of computing a flip point is proportional (with a constant factor in complexity) to the cost of evaluating the output of the model for that input. So, assuming that we want to audit a particular model that is already in use on a computer, that computer would be able to compute the flip point and the explanation report as well. If the auditor wants the closest flip points for an entire dataset, they can be computed in parallel.

See \citep{yousefzadeh2019thesis} for more details on defining and computing flip points for 2-class, multi-class and quantified output.

\subsection{Constrained flip points} \label{sec_define_flipb}

Suppose, for a particular input, we are interested in the influence of a single feature on the output of our model. If the feature has discrete values (e.g., ``owns home", ``rents", ``no fixed address"), then, as is well known, we simply evaluate the model for the same input but different values for that feature.
If the feature has continuous values, though, we might be interested in the smallest change in that feature that changes the decision of the model.
Then to compute this closest {\em constrained} flip point we solve the optimization problem of Section \ref{sec_define_flipa} allowing only that feature to vary. This is a 1-variable optimization problem that can be solved by standard algorithms such as bisection and other methods used for linesearch.

If we want to allow $k$ (continuous or discrete) features to vary, then we find the closest constrained flip point by solving   the optimization problem of Section \ref{sec_define_flipa} but with only these $k$ variables, keeping the other features constant.
We solve this problem using the same approaches discussed for computing unconstrained flip points.

Finally, if we allow all features to vary, we solve our original optimization problem, \dc{obtaining an unconstrained flip point}.

\da{
\subsection{Two notes on defining flip points} \label{sec_define_flipc}


\ra{Sometimes datasets have redundant features, e.g., features that are linearly dependent or features that are not related to outputs. Redundant features may not contribute to the predictive power of \dc{the} model, and including them in training may even lead to over-fitting \citep{tolocsi2011classification}. In our numerical examples, we show that excluding nearly linearly dependent features may improve the generalization of models. So, it can be helpful to study the dependencies prior to training.

Moreover, knowing the dependencies among the features can help in choosing meaningful subsets of features for computing constrained flip points. \dc{for example, ``income" and ``net worth" may be} correlated in a dataset.} If we choose to vary a subset of features that contains ``income" while holding ``net worth" constant, the constrained flip point might not be very meaningful.

So for many reasons, it can be desirable to identify \ra{the dependencies among the features in a dataset}. In our computational examples, we do this using the pivoted QR decomposition \cite[Chap. 5]{golub2012matrix} of a data matrix $\bfD$ whose rows are the training data points and whose columns are features. This decomposition reorders the columns, pushing linearly dependent columns (redundant features) to the right and forming
\[
\bfD \bfP = \bfQ \bfR,
\]
where $\bfP$ is the permutation matrix, $\bfQ$ has orthogonal columns, and $\bfR$ is zero below its main diagonal.
The degree of independence of the features can be determined by measuring the matrix condition number of \dc{leading principal submatrices} of $\bfR$, or by taking the matrix norm of trailing sets of columns. The numerical rank of $\bfD$ is the \dc{dimension of the largest leading principal submatrix of $\bfR$} with a sufficiently small condition number or, equivalently, the smallest number of leading columns that yields a small norm for the trailing columns.

Alternatively, the singular value decomposition (SVD) of $\bfD$ can be used in a similar way \cite[Chap. 2]{golub2012matrix}. In this case, the numerical rank is the number of sufficiently large singular values. The SVD will identify principal components \dc{(i.e., linear combinations of features in decreasing order of importance)}, and unimportant ones can be omitted. The most significant combinations of features can be used as training inputs, instead of the original features.
}

The underlying metric of these matrix decompositions is the Euclidean norm, so they are most easily justified for continuous features measured on a single scale, for example, pixel values in an image. For disparate features, the scale factors used by practitioners to define an appropriate norm for the optimization problem in Section \ref{sec_define_flipa} can be used to renormalize features before forming $\bfD$. \ra{Leaving the choice of scale factor to practitioners is suggested by \citet{spangher2018actionable} and \citet{wachter2018counterfactual}, too.}
}

\section{Using flip points to explain, audit and debug models} \label{sec_using}

\subsection{\ra{Individual-level auditing:} Providing \da{explanations and} feedback to users of a model}
\label{sec_feedback}

To generate a report like that in Figure \ref{fig_explain_report}, \da{we need to compute flip points and constrained flip points} in order to determine the {\em smallest} changes in the features that change the model's output.
\da{Algorithm \ref{alg_report} summarizes the use of constrained flip points in generating such a report, giving a user precise information on how individual features and combinations of features influenced  the model's recommendation for a given input.
 This has not previously been possible.}

\begin{algorithm}[h!]
\caption{Using constrained flip points to generate an explanation for a model's output for a specific input $\bfx$}
\label{alg_report}
\begin{flushleft}
\textbf{Given}: a trained model $\mathcal{N}$, a specific input $\bfx$,  and desired subsets of features to be investigated \\
\textbf{Produce}: an explanation report, giving various insights about the  model's output for  $\bfx$
\end{flushleft}
\begin{algorithmic}[1] 
\STATE Compute the closest flip point to $\bfx$
\STATE Compute constrained flip points for $\bfx$, allowing one feature to change at a time
\STATE Group the features that have the same measurement scale
and compute the constrained flip points for $\bfx$ in subspaces defined by each feature group
\STATE Compute the constrained flip point for $\bfx$ allowing any desired subset of features to change
\STATE Generate an explanation report based on the computed flip point and constrained flip points 
\end{algorithmic}
\end{algorithm}

\subsection{\ra{Group-level auditing:} Studying the behavior of a model towards groups of individuals}

\ra{It is important to audit and explain the behavior of models, not only on the individual-level, but also towards groups. Groups of interest can be an entire dataset or specific subsets within it, such as people with certain age, gender, education, etc. The information obtained from the group-level analysis can reveal systematic traits or biases in model's behavior. It can also reveal the role of individual features or combinations of features on the overall behavior of model.
}

\da{
Algorithm \ref{alg_directions} presents some of the ways that flip points can yield insight into these matters. 
}\ra{
By computing the closest flip points for a group of individuals, we obtain the vectors of directions to the decision boundary for them. We call \dy{these} directions {\em flip directions}. Using pivoted QR decomposition or principal component analysis (PCA) on the vectors of directions, we can identify important patterns and traits in \dc{a model's} decision making for the group of individuals under study.

For example, consider auditing a cancer prediction model for group of individuals with cancerous tumors. After computing the flip directions, we can study the patterns of change for that population, e.g., which features have changed most significantly and in which direction.

We can also study the effect of specific features on \dc{a} model's decision making for specific groups. For this type of analysis, we compute constrained flip points for the individuals in the group, allowing only the feature(s) of interest to change. We then study patterns in the directions of change. For example, when auditing a model trained to evaluate loan applications, we might examine the effect of age for people who have been \dc{denied}. We can compute constrained flip points for those individuals, allowing only the feature of age to change, and then study the patterns in flip directions, i.e., in which direction ``age" should change  and to what extent in order to change the decisions \dc{for that population}.
}

\ra{
\dc{We might also want} to examine the effect of gender for the same loan application model.} \da{To do this, \dc{we pair each data point with an identical one but of opposite gender.} We compute \dc{flip points} for all of the inputs and look for patterns: For the \dc{paired} points whose classification did not change, did the mean/median distance to the decision boundary change significantly? For the points whose classification changed, do the directions to the boundary have any commonalities, as revealed by pivoted QR or principal component analysis (PCA)?
}


\begin{algorithm}[h!]
\caption{Auditing a model's behavior on training or testing data}
\label{alg_directions}
\begin{flushleft}
\textbf{Given}: a trained model $\mathcal{N}$ and \ra{a data} matrix $\bfD$  \\
\textbf{Produce}: various insights into the behavior of the model
\end{flushleft}
\begin{algorithmic}[1] 
\STATE Compute the pivoted QR decomposition of $\bfD$ to identify redundant features. If appropriate, consider training a model with a smaller number of data features.
\STATE Compute the closest (or constrained) flip points for all the data in $\bfD$, forming a matrix $\bfB$. 
\STATE For correctly classified points (and then again for incorrectly classified ones), form $\bfF = \bfB - \bfD$, the matrix  of directions from data  points to flip points
\STATE Perform pivoted QR on $\bfF$ to identify features that are most and least influential in flipping the decisions of the model.
\STATE If $\bfF$ is close to rank deficient, then the set of directions to the decision boundary is of lower dimension than the number of features and \ra{it would be insightful to investigate the source of rank deficiency, i.e., zero columns and/or linearly dependent columns and their corresponding features.}
\STATE Compute the principal components of $\bfF$ to identify commonalities among the directions to the boundary from the training points.
\STATE \ra{Study the frequency of change between points and their flip points for each feature to gain insight about influence of features. Some features may change rarely among the population while some features may change frequently, indicating traits about the model.}
\STATE For a (binary) feature that should not affect the output classification, consider the dataset $\hat{\bfD}$ that has the opposite value for that feature. Compute  the resulting classifications. For points whose classification did not change, compute the mean change in distance to the boundary; ideally, this will be small. For points whose classification changed, pivoted QR or PCA analysis on the direction matrix will identify possible sources of the model's \ra{rationale}.
\end{algorithmic}
\end{algorithm}

\subsection{Debugging a model}

If we determine that the model's behavior is undesirable for a particular set of inputs, we would like to alter the decision boundaries to change that behavior. \ra{For example, when there is bias towards a certain feature, it usually means data points are close to decision boundaries in that feature dimension. By computing constrained flip points in that dimension, \dc{adding} them to the training set with the same label, and retraining, we can push the decision boundaries away from the inputs in that dimension. This tends to change the behavior of models, as we show in our numerical results.

Moving the decision boundaries away from the training data also \dy{tends} to improve the generalization of deep learning models as reported by \citet{elsayed2018large} and \citet{yousefzadeh2019interpreting}.

\rb{It is also possible to create flip points and teach them to the model with a flip label \dy{(i.e., $z_1 = z_2 = 1/2$)}, in order to define a decision boundary in certain locations.}
}

\section{Results} \label{sect_results}

\da{Here, we demonstrate our techniques for explaining, auditing, and debugging deep learning models on three different datasets \ra{with societal themes}. We use three software packages, NLopt \citep{nlopt}, IPOPT \citep{wachter2006implementation}, and the Optimization Toolbox of MATLAB, as well as our own custom-designed homotopy algorithm \citep{yousefzadeh2019interpreting}, to solve the optimization problems.} The algorithms almost always converge to the same point. The variety and abundance of global and local optimization algorithms in the above optimization packages give us confidence that we have indeed usually found the closest flip point.

For the two first examples, the FICO challenge and the Credit dataset, we compare our results with  \da{two} recent papers that have used those datasets. To make the comparison fair and easy, for each dataset we make the same choices about the data (such as cross validation, portion of testing set, etc.) as each of those papers.

\subsection{FICO Explainable ML Challenge} \label{sect_fico}

This dataset has 10,459 observations with 23 features, and each data point is labeled as ``Good" or ``Bad" risk. We randomly pick 20\% of the data as the testing set and keep the rest as the training set. We regard all features as continuous, since even ``months" can be measured that way. The description of features \da{is} provided in Appendix \ref{appx_ficofeatures}.

\subsubsection{\bf Eliminating redundant features.} The condition number of the matrix formed from the training set is 653. Pivoted QR factorization  finds that features ``MSinceMostRecentTradeOpen", ``NumTrades90Ever2DerogPubRec", and ``NumInqLast6Mexcl7days" are the most dependent columns; discarding them leads to a training set with condition number 59.
Using the data with 20 features, we train a neural network with 5 layers, achieving 72.90\% accuracy on the testing set. A similar network trained with all 23 features achieved 70.79\% accuracy, confirming the effectiveness of our decision to discard three features.

\subsubsection{\bf Individual-level explanations.} 
As an example, consider the first datapoint, corresponding to a person with ``Bad" risk performance. The feature values for this data point are provided in Appendix \ref{appx_ficofeatures}. \da{The closest (unconstrained) flip point is virtually identical to the data point except in five features, shown in Table \ref{table-fico}.}

\begin{table}[h]
\caption{Difference in features for data point \# 1 in the FICO dataset and its closest flip point.}
\label{table-fico}
\begin{center}
\begin{small}
\begin{tabular}{c >{\raggedleft\arraybackslash}p{0.5cm} >{\raggedleft\arraybackslash}p{1.1cm} >{\raggedleft\arraybackslash}p{1.1cm}}
\toprule
Feature & Input $\#1$ & Closest flip point (relaxed) & Closest flip point (integer) \\
\toprule
AverageMInFile & 84 & 105.6 & 111.2 \\
\midrule
NumSatisfactoryTrades  & 20 & 24.1 & 24 \\
\midrule
MSinceMostRecentDelq  & 2 & 0.6 & 0 \\
\midrule
NumTradesOpeninLast12M  & 1 & 1.7 & 2 \\
\midrule
NetFractionRevolvingBurden  & 33 & 19.4 & 8.5 \\
\bottomrule
\end{tabular}
\end{small}
\end{center}
\end{table}

\begin{figure}[h]
\centering
\includegraphics[width=1\columnwidth]{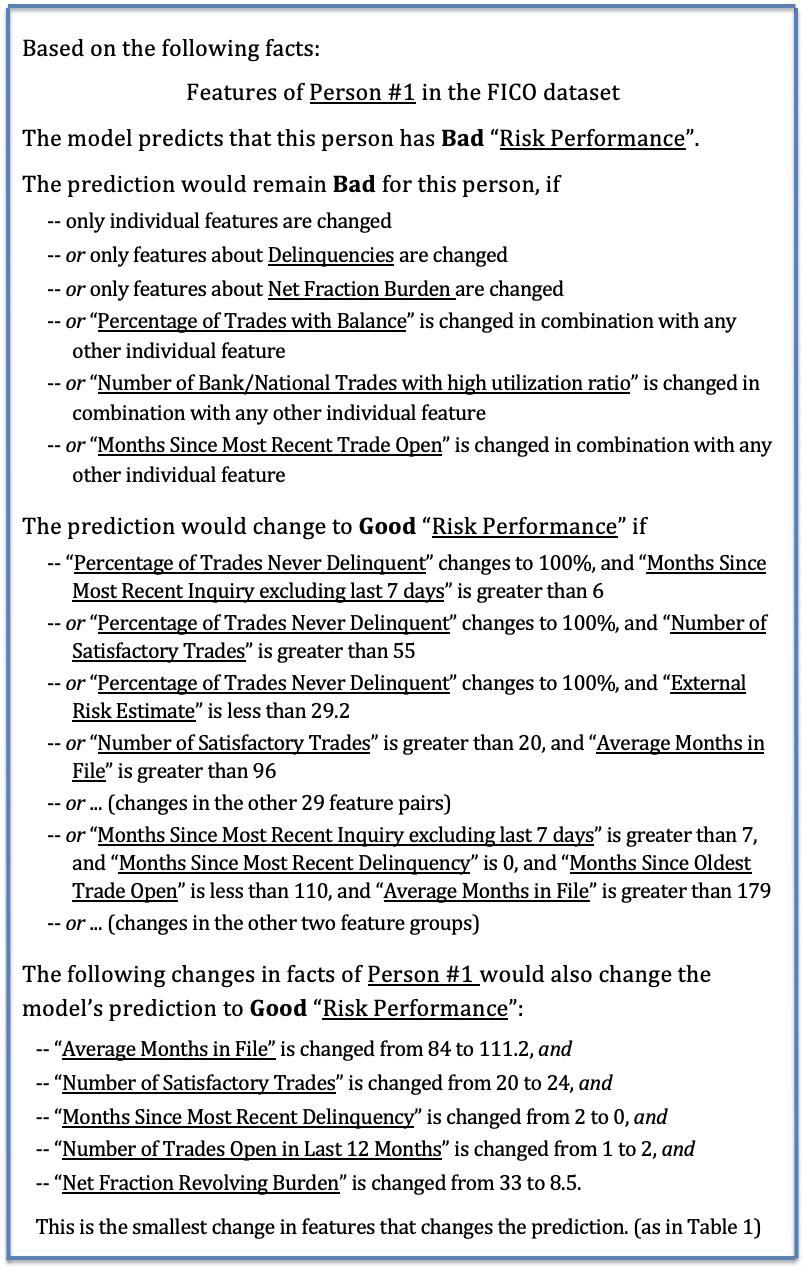}
\caption{A sample explanation report for data point \#1 in the FICO dataset, classified by a deep learning model.}
\label{fig_explain_report_fico}
\end{figure}

\da{Next}, we allow only a subset of \da{the} features to change \da{and compute  constrained flip points. We explore the following subspaces:}
\begin{enumerate}
	\item Only one feature is allowed to change at a time. \da{None} of the 20 features \da{is} individually capable of flipping the decision of the model.
	\item Pairs of features are allowed to change at a time. \da{Only} a few of the pairs (29 out of 190) can flip the output. 13 of these pairs involve the feature ``MSinceMostRecentInqexcl7days" as partially reflected in the  \da{explanation report of Figure \ref{fig_explain_report_fico}.} 
		\item Combinations of features that share the same measurement scale are allowed to change at a time. We have five distinct groups: features that are measured in ``percentage", ``number of months", ``number of trades", ``delinquency measure", and ``net fraction burden". The last two feature groups are not capable of flipping the prediction of the model by themselves. 
\end{enumerate}
The explanation summary report \da{resulting from these computations}  is shown in Figure \ref{fig_explain_report_fico}. The top two sections show the results of computing constrained flip points, first, points where \dy{no constrained flip point exists and} the label does not change, and then points with different label. The bottom section displays the unconstrained flip point.  This shows that the output of a deep learning model can be explained clearly and accurately to the user \da{to any desired level of detail}.  The answer to other specific questions can also be found easily by modifying the optimization problem.

We note that the time it takes to find each flip point is only a few milliseconds using a 2017 MacBook, hence this report can be generated in real-time.

%
%
%
%

\subsubsection{\bf Group-level explanations.} 
Using pivoted QR on the matrix of directions between data points labeled ``Bad" and their flip points, \da{we find that, individually,} the three most influential features are ``AverageMInFile", ``NumInqLast6M", and ``NumBank2 NatlTradesWHighUtilization". Similarly, for the directions that flip a ``Good" to a ``Bad", the three most influential features are ``AverageMInFile", ``NumInqLast6M", and ``NetFractionRevolvingBurden". In both cases, ``ExternalRiskEstimate" has no influence.

We perform PCA analysis on the subset of directions that flip a ``Bad" to ``Good" risk performance. The first principal component reveals that, for this model, the most prominent features with positive impact are ``PercentTradesNeverDelq" and ``PercentTradesWBalance", while the features with most negative impact are ``MaxDelqEver" and ``MSinceMostRecentDelq".
These conclusions are similar to the influential features reported by \cite{chen2018interpretable}, however, our method gives more detailed insights\ra{, since it includes an individual-level explanation report and also analysis of the group effects}.


\subsubsection{\bf Effects of redundant variables.}
Interestingly, for the model trained on all 23 features, the \da{three} most significant \da{individual} features in flipping its decisions are ``MSinceMostRecentTradeOpen", ``NumTrades90Ever2DerogPubRec" and ``NumInqLast6Mexcl7days", exactly the three dependent features that we discarded for the reduced model.
\da{Thus,} the decision of the trained model is more susceptible to changes in the dependent features, compared to changes in the independent features.

\ra{This reveals an important vulnerability of machine learning models regarding their training sets. For this dataset, when dependent features are included in the training set, the accuracy on the training set remains the same, but it adversely affects the accuracy on the testing set, i.e., generalization. Additionally, when those redundant features are included, they become the most influential features in flipping the decisions of the model, making the model vulnerable.
}

\subsubsection{\bf Auditing the model using flip directions.}
\begin{figure}[h]
\centering
\includegraphics[width=0.99\columnwidth]{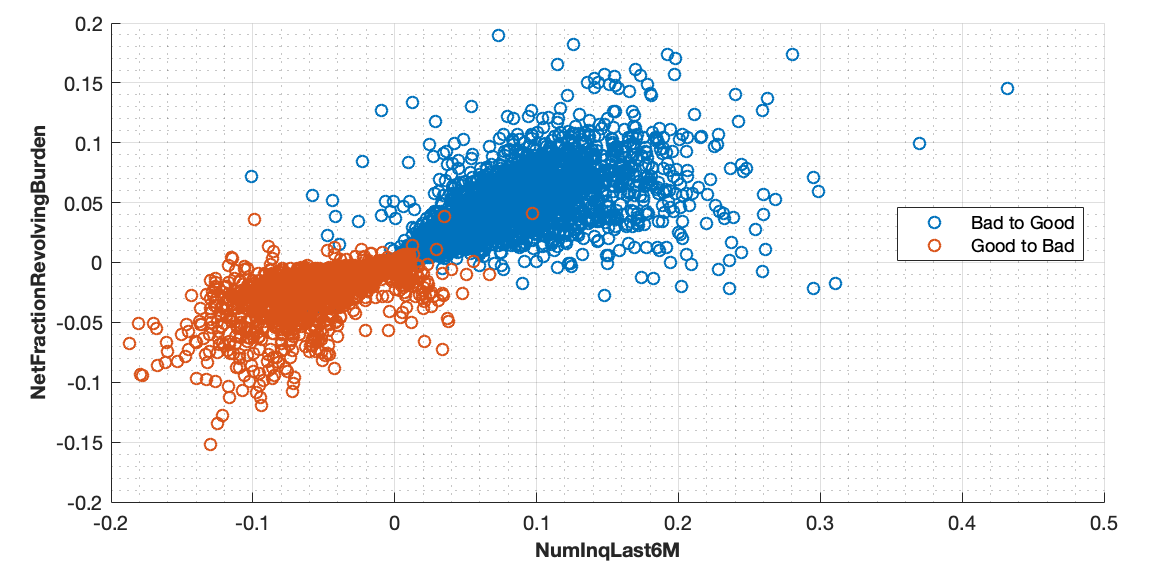}
\caption{Directions between the inputs and their closest flip point for two influential features. Points are distinctly clustered based on the direction of the flip.
}
\label{fig_fico_directions}
\end{figure}

Figure \ref{fig_fico_directions} shows the directions of change to move from the inputs to the closest flip points for features ``NumInqLast6M" and ``NetFractionRevolvingBurden"\ra{, which are the most influential features given by the pivoted QR algorithm}. \ra{Even though flip points are unconstrained, directions of change for these two features} are distinctly clustered for flipping a ``Bad" label to ``Good" and vice versa. 

\begin{figure}[h]
\centering
\includegraphics[width=0.99\columnwidth]{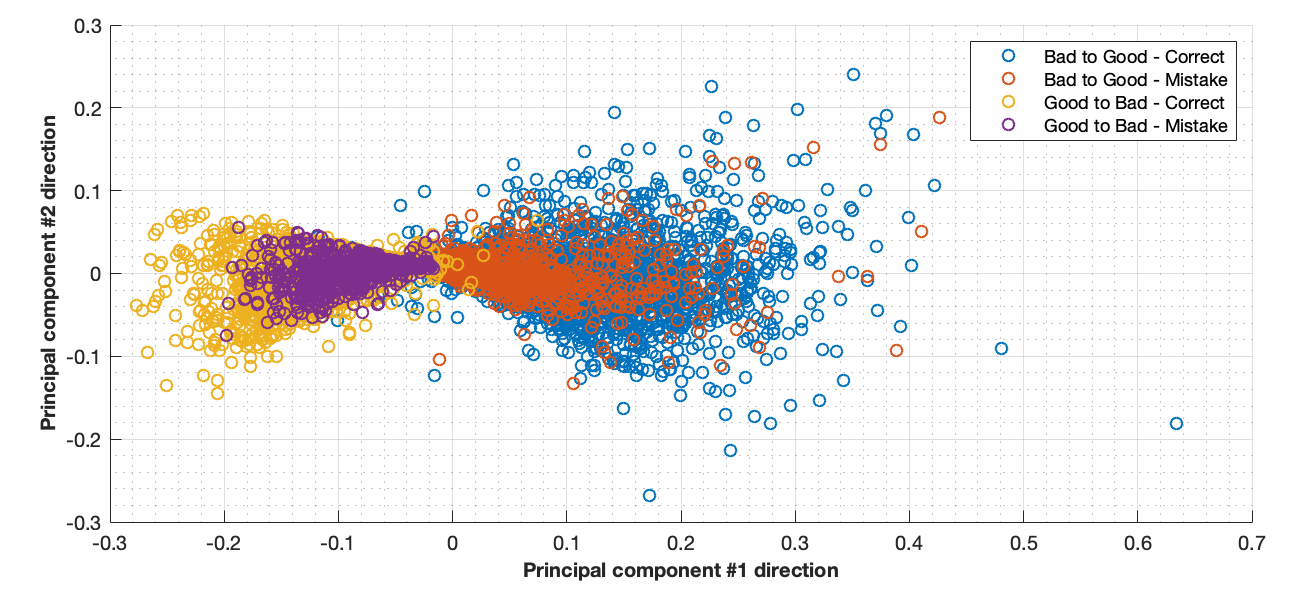}
\caption{Change between the inputs and their unconstrained flip points in the first two principal components. Directions are clustered into two convex cones, exactly in opposite directions.} 
\label{fig_fico_pca}
\end{figure}

Furthermore, Figure \ref{fig_fico_pca} shows the directions in coordinates of the first two principal components. 
We can see that the flip directions are clearly clustered into two convex cones, exactly in opposite directions. Also, we see \da{that} misclassified inputs are relatively close to their \da{flip points} while correct predictions can be close or far. 
\da

\ra{
\subsubsection{{\bf Comparison}}
The interpretable model developed by \citet{chen2018interpretable} reports the most influential features which are similar to our findings above, e.g., ``PercentTradesNeverDelq" and ``AverageMInFile". However, their model is inherently interpretable, and their auditing \dc{method} is not applicable to deep learning models. They also do not provide an explanation report on the individual-level, like the one we provided in Figure~\ref{fig_explain_report_fico}. 

We note that our goal, here, is to show how a deep learning model utilized for this application can be audited. We do not necessarily advocate for use of deep learning models over other models.}

\subsection{Default of credit card clients }
This  dataset from the UCI Machine Learning Repository \citep{Dua2017} has 30,000 observations, 24 features, and a binary label predicting whether or not the person will default on the next payment. 

We binarize the categorical variables ``Gender", ``Education", and ``Marital status"; \da{the categories that are active for a data point have binary value of 1 in their corresponding features, while the other features are set to zero. When searching for a flip point, we allow exactly one binary feature to be equal to 1 for each of the categorical variables. } The condition number of the training set is 129 which implies linear independence of features. Using a 10-fold cross validation on the data, we train a neural network with 5 layers (details in Appendix \ref{appx_models}), to achieve accuracy of 81.8\% on the testing set, \da{slightly} higher than the accuracy of around 80.6\% reported by \cite{spangher2018actionable}. When calculating the closest flip points, we require the categorical variables to remain discrete.

\subsubsection{\bf Individual-level explanations}

\begin{figure}[h]
\centering
\includegraphics[width=1\columnwidth]{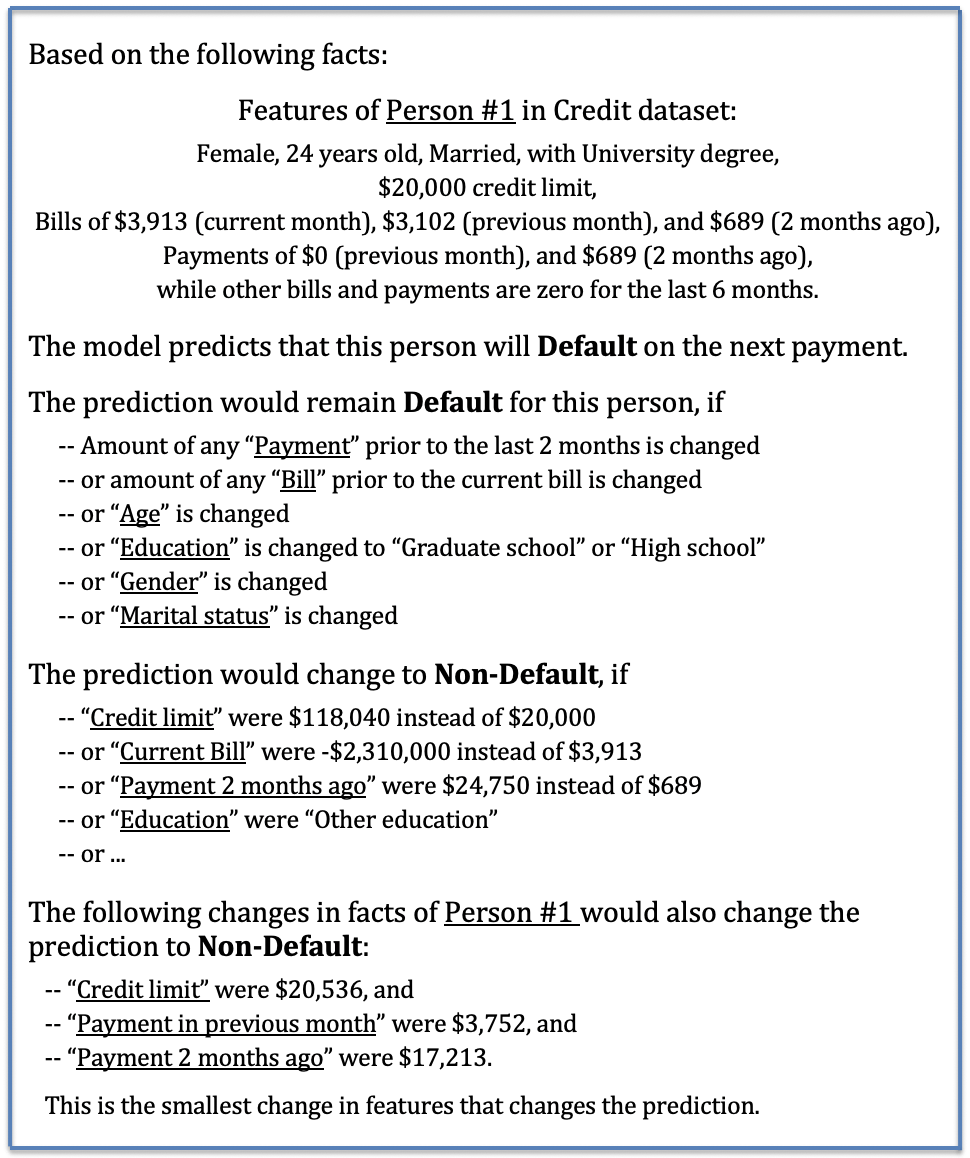}
\caption{A sample explanation report for data point \#1 in the Credit dataset, predicted to default on the next payment. The deep learning predicts the labels for the testing data, well. But, what it takes to change the prediction of model sometimes does not seem rational.}
\label{fig_explain_report_credit}
\end{figure}

We consider the data point \#1 in this dataset which is classified as ``default", and \da{compose the explanation report shown in Figure \ref{fig_explain_report_credit}}. When we examine the effect of features, we see that \da{any of} 4 features can flip the prediction of the model, individually.

When examining this report for input \#1, we find some flaws in the model. For example, in order to flip the prediction of the model to non-default, one option is to reduce the amount of the current bill to -\$2,310,000, while reducing the bill to any number larger than that would not flip the prediction.

\rb{Requiring any negative balance on the bill is irrational, because as long as the bill is zero, there would be no chance of default. In fact, one would expect the prediction of non-default if the current bill is changed to zero, for any datapoint. But, the training set does not include such examples, and clearly, our model has not learned such an axiom.}
Requiring the large payment of \$24,750 (for 2 months ago) in order to flip the prediction seems questionable, too, considering that the current bill is \$3,913. 

Therefore, despite the model's good accuracy on the testing data, the explanation for its prediction \da{reveals} flaws in its behavior for data point \#1. These flaws  would not have been \dc{noticed} without investigating \da{the} decision boundaries. Fortunately, because of our auditing, we know that the model needs to be improved before it is deployed.


\subsubsection{\bf Group-level auditing using flip points.} 

\da{Examining the flip points for the training data reveals model characteristics that should be understood by the users. Here is one example.}

  Gender \da{does not have much influence in the decisions of the model,} as only about 0.5\% of inputs have a different gender than their flip points. Hence, gender is not an influential feature for this model. This kind of analysis can be performed for all the features, in more detail.
  



\subsubsection{\bf Group-level auditing using flip directions.}  We perform pivoted QR decomposition on the directions to the closest flip points. The results show that ``BILL-AMT3"\footnote{``BILL-AMTx" stands for the Amount of Bill in \$, x month(s) ago.} and ``BILL-AMT5" are the most influential features, and ``Age" has the least influence in \da{changing} the predictions. In fact, there is no significant change between the age of \da{any of} the inputs and their closest flip points.

\subsubsection{\bf Debugging the model using flip points.}
In both our training and testing sets, about 52\% of individuals have age less than 35.
\da{Following \cite{spangher2018actionable}, we remove 70\% of the young individuals from the training set, so that they are under-sampled.}  We keep the testing set as before and obtain 80.83\% accuracy on the original testing set. We  observe that \dc{now,} ``Age" is the 3$^{\text{rd}}$ most influential feature in flipping its decisions. Moreover, PCA analysis shows that \da{lower} Age has a negative impact on the ``no default" prediction and vice versa.

We consider all the data points in the training set labelled as ``default" that have closest flip point with older age, and all the points labelled ``no default" that have closest flip point with younger age. We add all those flip points to the training set, with the same label as their corresponding data point, and train a new model. \da{Now} Age has become the 11$^{\text{th}}$ \da{most} influential feature and it is no longer significant in the first principal component of \da{the flip directions}; hence, the bias against Age has been reduced. \ra{Also, testing accuracy slightly increases to 80.9\%.} 

Adding synthetic data to the training set has great potential to change the behavior of \dc{a} model, but we cannot rule out unintended consequences.
By investigating the influential features and PCA analysis, we see that the model has been altered only with respect to the Age feature, and the overall behavior of model has not changed.

\ra{
\subsubsection{Comparison}
\citet{spangher2018actionable} has used a logistic regression model for this dataset, achieving 80.6\% accuracy on testing, less than our 81.8\% accuracy. Their method for computing flip points is limited to linear models and not applicable to deep learning. They also do not provide an explanation report like the one in Figure~\ref{fig_explain_report_credit}.

They have reported that under-sampling young individuals from the training set makes their model biased towards young age, similar to ours. However, they do not use flip points to reduce the bias, which we successfully did.
}

\subsection{Adult Income dataset}
\rb{
The Adult dataset from the UCI Machine Learning Repository \cite{Dua2017} has a combination of discrete and continuous variables.  Each of the  32,561 data points in the training set and 16,281 in the testing set are labeled, indicating whether the individual's income is greater than 50K annually.
%
%
There are 6 continuous variables including Age, Years of education, Capital-gain, Capital-loss, and Hours-per-week of work. We binarize the discrete variables: Work-class, Marital status, Occupation, Relationship, Race, Gender, and Native country.  
Our trained model \da{considers 88 features and} achieves accuracy 86.08\% on the testing sets comparable to best results in the literature \citep{Dua2017}. 
Our aim here is to show how a trained model can be audited.

\subsubsection{\bf Individual-level auditing.} 


As an example, consider the first data point in the testing set, corresponding to a \dy{25-year-old} Black Male, with 11th grade education and native country of United States, working 40 hours per week in \dy{the} Private sector as Machine-operator-inspector and income ``$\le 50$K", correctly classified by the model. He has never married and \dy{has} a child/children.

We compute the closest flip point for this individual, allowing all the features to change. Table~\ref{table_adult_individual} shows the features that have changed for this person in order to flip the model's classification for him to the high income bracket. Other features such as gender, race, work-class \dy{have} not changed and are not shown in the table. Directions of change in the features \dy{are} generally sensible: e.g., working more hours, getting a higher education, working in the Tech-sector, and being older generally have a direct relationship with higher income. Being married instead of being a single parent is also known to have a relationship with higher income.

\begin{table}[h]
\caption{Difference in features for Adult dataset testing point $\#1$ and its closest flip point.}
\label{table_adult_individual}
\begin{center}
\small
\begin{tabular}{  p{2.2cm} p{2.4cm}  p{2.5cm}   }
\toprule
Data & Input $\#1$ in testing set & Closest flip point  \\
\midrule
Age    & 25 & 30.3  \\
\midrule
Years of education & 7 (11th grade) & 8 (12th grade)  \\
\midrule
Marital status  & Never-married & Married-ArmedForcesSpouse  \\
\midrule
Relationship   & Own-child & Husband  \\
\midrule
Occupation  & Machine-operation-inspection & Tech-support  \\
\midrule
Hours-per-week of work  & 40 & 41.8  \\
\bottomrule
\end{tabular}
\end{center}
\end{table}

We further observe that none of the features \dy{individually} can flip the classification, but  \dy{certain} constrained flip points  \dy{can} provide additional insights about the behavior of the model. 

Let's consider the effect of race. \dy{The softmax score for this individual is $0.9989$} for income ``$\le 50$K". Changing the race does not affect the softmax score more than 0.0007. This observation about softmax score might lead one to believe that the model is neutral about race, at least for this individual. However, that would not be completely accurate in all circumstances, as we will explain. If we keep all features of this individual the same and only change his race to Asian, the closest flip point for him would be the same as before, except for Age of 29.9 and Hours-per-week of 42.3. The differences in flip points for the Black and Asian are not large enough to draw a conclusion.

Let's now take one step further and fix his education to remain 11th grade and re-examine the effect of race. The resulting closest flip points are shown in Table~\ref{table_adult_individual_race} for two cases: where his race is kept Black and where it is changed to Asian. Clearly, being Asian requires considerably smaller changes in other features in order to reach the decision boundary of \dy{the} model and flip to the high income class. This shows that race can be an influential feature in model's classifications of people with low education. Having education above the 12th grade for this individual makes the effect of race negligible.

\begin{table}[h]
\caption{Race can be an influential feature for individuals with low education. Closest flip points for testing point $\#1$ in Adult dataset when education is fixed to 7th grade and race is changed from Black to Asian.}
\label{table_adult_individual_race}
\begin{center}
\small
\begin{tabular}{  p{2.2cm} p{2.4cm}  p{2.5cm}   }
\toprule
Data & Closest flip point (Black) & Closest flip point (Asian)  \\
\midrule
Age    & 41.9 & 32.4  \\
\midrule
Years of education & 7 (11th grade) & 7 (11th grade)  \\
\midrule
Marital status  & Married-ArmedForcesSpouse & Married-ArmedForcesSpouse  \\
\midrule
Relationship   & Husband & Husband  \\
\midrule
Occupation  & Tech-support  & Tech-support  \\
\midrule
Hours-per-week of work  & 44.3 & 42.4 \\
\bottomrule
\end{tabular}
\end{center}
\end{table}

We further observe that gender does not have an effect on \dy{the model's} classification for this individual, whether the education is high or low. \dy{The} effect of other features related to occupation and family can also be \dy{studied.}




\subsubsection{\bf Group-level auditing using flip points.}

As an example, we consider the group of people with native country of Mexico. About 95\% of this population have income ``$\le 50$K" and 77\% of them are Male. We compute the closest flip point for this population and investigate the patterns in them and how frequently features have changed from data points to flip points, and in what way.

Let's consider the effect of gender. 99\% of the females in this group have income ``$\le 50$K" and for 40\% of them, their closest flip point is Male. Among the Males, however, less than 1\% have a Female flip point; some of \dy{these are} high-income individuals \dy{for whom the change in gender flips them}  to low-income.

Let's now consider \dy{the patterns in flip points that change low income males and females to}  high-income. For occupation, the most common change is entering into the Tech-sector and the most common exit is from the Farming-fishing occupation. For relationship, the most common change is to being married and the most common exit is from being Not-in-family and Never-married. 
Among the continuous features, Years of education and Capital-gain have changed most frequently.

\subsubsection{\bf Group-level auditing using flip directions.} 
%
Consider the subset of directions that flip a ``$\le 50$K" income to ``$>50$K" for the population with native country of Mexico. The first principal component reveals that, for this model and this population, the most prominent features with positive impact are having a higher education, having Capital-gain, and working in the Tech-sector, while the features with most negative impact are being Never-married, being Female, and having Capital-loss. Looking more deeply at the data, pivoted QR decomposition of the matrix of flip directions reveals that some features, such as being Black and native country of Peru have no impact on this flip.

\subsubsection{\bf Group-level analysis of flip directions for misclassifications.} 
Besides studying specific groups of individuals, we can also study the misclassifications of the model. PCA on the \dc{flip directions for all the misclassified points} in the training set shows that Age has the largest coefficient in the first principal component, followed by Hours-per-week of work. The most significant feature with negative coefficient is having Capital-gain. These features can be considered the most influential in confusing and de-confusing the model. PCA on the \da{flip directions} explains how our model is influenced by various features and its vulnerabilities for misclassification. It thus enables us to \dc{create} inputs that are mistakenly classified for adversarial purposes, \dy{as} explained by \citet{lakkaraju2019fool} and \citet{slack2019can}.

}

\section{Comparison with other interpretation approaches for deep learning} \label{sec_compare}
Our use of flip points for interpretation and debugging \dd{builds on existing methods in the literature but provides more comprehensive capabilities}. For example, \citet{spangher2018actionable} \dc{compute} flip sets only for linear classifiers and \dc{do} not use them to explain the overall behavior of the model, identify influential features, or debug.

LIME \citep{ribeiro2016should} and Anchors \citep{anchors:aaai18} rely on sampling around an input in order to investigate decision boundaries, inefficient and less accurate than our approach, and the authors do not propose using their results as we do. \rb{LIME provides a coefficient for each feature (representing a hyperplane) which may not be easily understandable by non-experts (e.g., a loan applicant or a clinician), especially when dealing with a combination of discrete and continuous features. LIME's approach also relies on simplifying assumptions, \dd{such as the ability to approximate decision boundaries by hyperplanes,} which leads to contradictions between the LIME output and the model output \cite{white2019measurable}, a.k.a. infidelity. So, our method has an accuracy advantage over their method, too. Moreover, their reliance on random perturbations of data points can be considered a computational limitation when applying their method to deep learning models.}


%

The \dd{interpretation we provide for nonlinear deep learning models is comparable in quality and} \rb{extent to the interpretations} provided in the literature for simple models. For example, the model suggested by \cite{chen2018interpretable} for the FICO Explainable ML dataset reports the most influential features in decision making of their model, similar to our findings in Section \ref{sect_fico}, and investigates the overall behavior of the model, similar to our results for the Adult dataset. But, \rb{their methods are not applicable for auditing deep learning models. Moreover, they do not provide a detailed explanation report.}

We also show how decision boundaries can be altered to change the behavior of models, an approach not explored for deep learning models.




\section{Conclusions and future work} \label{sec_conclusion}

We have proposed the computation of flip points in order to \da{explain,} debug, and audit deep learning models with continuous output.
We demonstrated that computation of the closest flip point for an input to a continuous model provides useful information to the user, explaining why a model produced a particular output and identifying any small changes in the input that would change the output.
Flip points also provide useful information to model auditors, exposing bias and revealing patterns in misclassifications. We provide an algorithm to formalize the auditing procedure.
Finally, model developers can use flip points in order to alter the decision boundaries and eliminate undesirable behavior of a model.

Our proposed method has accuracy advantages over existing methods in the literature, and it also has practical advantages such as fast interpretation for individual inputs and the ability to communicate with non-expert users (such a loan applicant or a clinician) via an explanation report.

For future work, we would consider models with continuous outputs \da{other than} classification models, for example, a model that recommends the dose of a drug for patients. Other directions of research \da{include} auditing image classification models, \da{expanding on work in \citep{yousefzadeh2019interpreting}}, and text analysis models that have a societal impact.  \da{Our methods can} promote fairness, accountability and transparency \da{in deep learning models}.

\clearpage
\appendix

\setcounter{table}{0}
\renewcommand{\thetable}{A\arabic{table}}
\renewcommand{\thealgorithm}{A\arabic{algorithm}}

\section{Description of variables for the FICO dataset} \label{appx_ficofeatures}

\begin{table}[h!]
\setlength{\tabcolsep}{5pt}
\caption{Variable descriptions for the FICO dataset.}
\label{table_ficofeatures}
\begin{center}
\begin{footnotesize}
\begin{sc}
\begin{tabular}{  p{3.5cm} | p{2.9cm} | p{.5cm} }
\toprule
Variable name & Description & data point \#1   \\
\midrule
ExternalRiskEstimate & Consolidated version of risk markers   & 55  \\
\midrule
MSinceOldestTradeOpen  & Months Since Oldest Trade Open  &  144 \\
\midrule
MSinceMostRecentTradeOpen & Months Since Most Recent Trade Open  &  4 \\
\midrule
AverageMInFile & Average Months in File  &  84 \\ 
\midrule
NumSatisfactoryTrades & Number of Satisfactory Trades   &  20  \\
\midrule
NumTrades60Ever2DerogPubRec &  Number of Trades 60+ Ever  & 3 \\
\midrule
NumTrades90Ever2DerogPubRec &  Number of Trades 90+ Ever  & 0  \\
\midrule
PercentTradesNeverDelq &  Percentage of Trades Never Delinquent  & 83 \\
\midrule
MSinceMostRecentDelq &  Months Since Most Recent Delinquency  & 2 \\
\midrule
MaxDelq2PublicRecLast12M &  Max Delinquency/Public Records in the Last 12 Months  &  3 \\
\midrule
MaxDelqEver &  Max Delinquency Ever  & 5 \\
\midrule
NumTotalTrades &  Number of Total Trades (total number of credit accounts)  &  23  \\
\midrule
NumTradesOpeninLast12M &  Number of Trades Open in the Last 12 Months  &  1  \\
\midrule
PercentInstallTrades &  Percentage of Installment Trades  & 43  \\
\midrule
MSinceMostRecentInqexcl7days &  Months Since Most Recent Inquiry (excluding last 7 days)  & 0 \\
\midrule
NumInqLast6M &  Number of Inquiries in the Last 6 Months  & 0 \\
\midrule
NumInqLast6Mexcl7days &  Number of Inquiries in the Last 6 Months (excluding last 7 days)  & 0 \\
\midrule
NetFractionRevolvingBurden &  Net Fraction Revolving Burden  & 33 \\
\midrule
NetFractionInstallBurden &  Net Fraction Installment Burden  & -8 \\
\midrule
NumRevolvingTradesWBalance &  Number of Revolving Trades with Balance  & 8 \\
\midrule
NumInstallTradesWBalance &  Number of Installment Trades with Balance  & 1 \\
\midrule
NumBank2NatlTradesWHighUtil & Number of Bank/National Trades with High Utilization Ratio  & 1  \\
\midrule
PercentTradesWBalance &  Percentage of Trades with Balance  &  69 \\
\bottomrule
\end{tabular}
\end{sc}
\end{footnotesize}
\end{center}
\end{table}

The name of each variable for the FICO dataset can be viewed in the first column of Table~\ref{table_ficofeatures}. The second column shows the corresponding description for each variable as provided by FICO. Additionally, the third column of this table shows the value of each variable for data point \#1. Detailed information about the challenge can be found here: \url{https://community.fico.com/s/explainable-machine-learning-challenge}.

\section{Code}  \label{appx_code}
The code along with a readme file and an example procedure are available at \url{https://github.com/roozbeh-yz/auditing}.

\setcounter{table}{0}
\renewcommand{\thetable}{C\arabic{table}}

\section{Information about the models} \label{appx_models}

Here, we provide more information about the models we have trained and used in Section \ref{sect_results}. 
We have used fully connected feed-forward neural networks with \da{up to} 6 hidden layers. The number of nodes for the models used for each data set is shown in Table \ref{table_nodes}. The activation function we have used in the nodes is the error function, as defined in \citep{yousefzadeh2019interpreting}. We have also used $\softmax$ on the output layer, and cross entropy for the loss function.

Models are designed using the method described by \citet{yousefzadeh2019refining}.

\begin{table}[h!]
\setlength{\tabcolsep}{5pt}
\caption{Number of nodes in neural network used for each data set.}
\label{table_nodes}
\vskip 0.15in
\begin{center}
\begin{small}
\begin{sc}
\begin{tabular}{  r | rrr }
\toprule
Data set & FICO & Credit & Adult   \\
\midrule
Input layer & 20 & 28 & 88  \\
\midrule
Layer 1 &13 & 14 & 40  \\
\midrule
Layer 2 & 9 & 9 & 32  \\
\midrule
Layer 3 & 6 & 8 & 24  \\
\midrule
Layer 4 & 5 & 8 & 20  \\
\midrule
Layer 5 & 4 & 7 & 16  \\
\midrule
Layer 6 & - & - & 14  \\
\midrule
Output layer & 2 & 2 & 2  \\
\bottomrule
\end{tabular}
\end{sc}
\end{small}
\end{center}
\vskip -0.1in
\end{table}

%
%
%
%
%
%
%
%

\clearpage
\bibliographystyle{ACM-Reference-Format}
\bibliography{Refs}

\end{document}